\documentclass[10pt,twocolumn,letterpaper]{article}

\usepackage{cvpr}
\usepackage{times}
\usepackage{epsfig}
\usepackage{graphicx}
\usepackage{amsmath}
\usepackage{amssymb}

\usepackage[pagebackref=true,breaklinks=true,letterpaper=true,colorlinks,bookmarks=false]{hyperref}

\cvprfinalcopy 

\def\cvprPaperID{****} 

\ifcvprfinal\fi
\begin{document}

\title{I Know How You Feel: Emotion Recognition with Facial Landmarks}

\author{Ivona Tautkute$^{1,2}$, Tomasz Trzcinski$^{1,3}$ and Adam Bielski$^1$\\
$^1$Tooploox $^2$Polish-Japanese Academy of Information Technology $^3$Warsaw University of Technology\\
{\tt\small \{firstname.lastname\}@tooploox.com}}

\maketitle

\begin{abstract}
Classification of human emotions remains an important and challenging task for many computer vision algorithms, especially in the era of humanoid robots which coexist with humans in their everyday life. Currently proposed methods for emotion recognition solve this task using multi-layered convolutional networks that do not explicitly infer any facial features in the classification phase. In this work, we postulate a fundamentally different approach to solve emotion recognition task that relies on incorporating facial landmarks as a part of the classification loss function. To that end, we extend a recently proposed Deep Alignment Network (DAN), that achieves state-of-the-art results in the recent facial landmark recognition challenge, with a term related to facial features. Thanks to this simple modification, our model called EmotionalDAN is able to outperform state-of-the-art emotion classification methods on two challenging benchmark dataset by up to 5\%.
\end{abstract}

\vspace{-0.3cm}
\section{Introduction}

Since autonomous AI systems, such as anthropomorphic robots, start to rapidly enter our lives, their ability to understand social and emotional context of many everyday situations becomes increasingly important. 
One key element that allows the machines to infer this context is their ability to correctly identify human emotions, such as happiness or sorrow. This is a highly challenging task, as people express their emotions in a multitude ways, depending on their personal characteristics, {\it e.g.} people with an introvert character tend to be more secretive about their emotions, while extroverts show them more openly. Although some simplifications can be applied, for instance reducing the space of recognized emotions, there is an intrinsic difficulty embedded in the problem of human emotion classification. 

Most of the currently available methods that address this problem use some variation of a deep neural network with convolutional layers. For instance~\cite{CNN_5} proposes to use a standard architecture of a convolutional neural network (CNN) with two convolutional, two subsamping and one fully connected layer. Before being processed, the image is spatially normalized with a pre-processing step. Another method presented in~\cite{baseline_cnn_inception} incorporates additional inception layers into the architecture, inspired by the Inception model~\cite{baseline_inception} that achieves state-of-the-art object classification results on the ImageNet dataset~\cite{imagenet}. Another variation of the Inception~\cite{baseline_inception}, uses the Inception-V3 model pretrained on the ImageNet dataset with a custom softmax layer trained specifically to classify emotions. Finally, the most recent method called EmotionNet~\cite{emotionet} and its extension EmotionNet2~\cite{emotionnet2} builds up on the ultra-deep ResNet architecture~\cite{DBLP:journals/corr/HeZRS15} and improves the accuracy by using face detection algorithm that reduces the variance caused by a background noise.

Although all the above methods rely on the state-of-the-art deep learning architectures, they draw their inspiration mostly from the analogical models that are successfully used for object classification tasks. We believe that as a result these approaches do not exploit intrinsic characteristics of how humans express emotions, {\it i.e.} by modifying their face expression through moving the landmark features of their faces \cite{sina}. We therefore propose to use a state-of-the-art facial landmark detection model -- Deep Alignment Network (DAN)~\cite{dan} -- and extend it by adding a surrogate term that aims to correctly classify emotions to the neural network loss function. This simple modification allows our method, dubbed EmotionalDAN, to exploit the location of facial landmarks and incorporate this information into the classification process. By training both terms jointly, we obtain state-of-the-art results on two challenging datasets for facial emotion recognition: CK+~\cite{ck} and ISED~\cite{ised}. 

\section{EmotionalDAN}

Our approach builds up on the Deep Alignment Network architecture~\cite{dan}, proposed initially for robust face alignment. The main advantage of DAN over the competing face alignment methods comes from an iterative process of adjusting the locations of facial landmarks. The iterations are incorporated into the neural network architecture, as the information about the landmark locations detected in the previous stage (layer) are transferred to the next stages through the use of facial landmark heatmaps. As a result and contrary to the competing methods, DAN can therefore handle entire face images and not patches which leads to a significant reduction in head pose variance and improves its performance on a landmark recognition task. DAN ranked $3^{rd}$ in a recent face landmark recognition challenge Menpo~\cite{menpo}.

In this work, we hypothesize that DAN's ability to handle images with large variation and provide robust information about facial landmarks transfers well to the task of emotion recognition. To that end, we extend the network learning task with an additional goal of estimating expressed facial emotions. We incarnate this idea by modifying the loss function with a surrogate term that addresses specifically emotion recognition task and we minimize both landmark location and emotion recognition terms jointly. The resulting loss function $\mathcal{L}$ can be therefore expressed as:

\begin{equation*}
\label{loss}
\mathcal{L} = \alpha \cdot \frac{\parallel S - S^{*} \parallel }{d} + \beta \cdot CE (E, E^{*} ) ,
\end{equation*}
\noindent where $S$ is the transformed output of predicted facial landmarks using Landmark Transform and Image transform layers~\cite{dan}, $E$ is the softmax output for emotion prediction.  $S^{*}$ is the vector of ground truth landmark locations, $d$ is the distance between the pupils of ground truth that serves as a normalization scalar and $E^{*}$ is the ground truth for emotion labels. $CE$ denotes Cross-Entropy loss. We weigh the influence of the terms with $\alpha$ and $\beta$ coefficients and after an initial set of experiments we fix their values to $\alpha=0.4$ and $\beta=0.6$.

\section{Experiments}
To evaluate the performance of the proposed EmotionalDAN method, we compute classification accuracy for the emotion recognition task, using several benchmark datasets, described in the next section. As our baselines, we use methods described in the introduction of this work, namely Convolutional Neural Network (CNN)~\cite{CNN_5} with 2 and 5 convolutional layers, Inception-V3~\cite{baseline_cnn_inception} and EmotionNet 2~\cite{emotionnet2}. 

When available, we use original implementations of the competing methods. For methods that are not made public, we implement them in Keras. Our EmotionalDAN is based on Tensorflow implementation of DAN~\cite{dan}.

\subsection{Datasets}
To train all the evaluated methods we use AffectNet~\cite{affectnet} - the largest available database for facial expression that contains over 1,000,000 face images collected from the Internet through emotion keyword querying. 
About half of the retrieved images were manually annotated for the presence of seven main facial expressions and 68 facial landmarks locations. This part of the dataset is used to train our methods. 

For testing, we use CK+~\cite{ck}, JAFFE~\cite{jaffe} and~\cite{ised} datasets with face images of over 180 individuals of different genders and ethnic background.

To provide a more holistic comparison of the methods, we split the emotions annotated in the test datasets into two scales: seven-grade scale with happy, sad, angry, surprised, disgust, fear  and neutral emotions, and a three-grade scale with positive, negative and neutral emotions. This way we obtain two complimentary evaluation sets with various amount of bias introduced by confusion of labelers and other confounding factors.






\subsection{Results}
Tables~\ref{results-accuracy-7} and \ref{results-accuracy-3} show the results of the evaluation of our EmotionalDAN method and the competing approaches. Although the accuracy varies between the tested datasets, our approach outperforms the competitors by a large factor of up to 5\% on two out of three benchmark datasets, namely on CK+ and ISED. The performance of our method is inferior to convolutional neural networks on the JAFFE dataset, although the accuracy values obtained on this dataset are generally lower than the competitors. We believe that this may be the result of a more challenging image acquisition conditions. Furthermore, our results show that convolutional neural networks achieve competitive results when compared with other methods despite their simplistic architecture. 
\vspace{-0.1cm}
\section{Application}
\begin{figure}[t!]
\begin{center}
\includegraphics[width=0.5\textwidth]{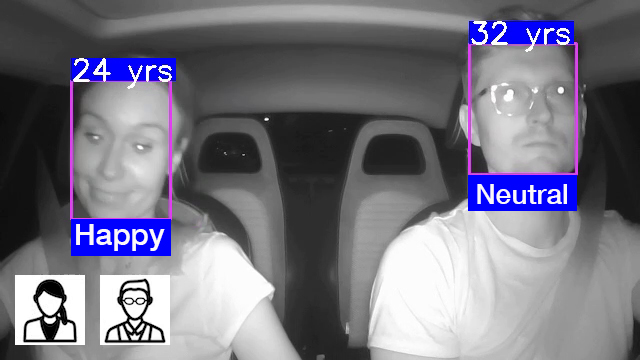}
\end{center}
   \caption{Our emotion recognition model in passenger detection system for autonomous cars. Emotion recognition is performed on detected facial regions}
\label{fig:car}
\end{figure}

We implement our emotion recognition model as a part of the in-car analytics system to be deployed in autonomous cars. Figure~\ref{fig:car} shows the results obtained by the camera installed inside a car. As autonomous car operation can potentially be influenced by emotions of the passengers ({\it e.g.} fear of speed expressed on passenger's face could signal the need for speed reduction), this is an excellent playground for our method to show its full potential. Although alternative applications are possible, we believe that this use case showcases the capabilities of our method and can serve as an interesting input to the driving system, typically focused on the exterior views from outside the car. 

\section{Conclusion}

In this paper, we overview a work-in-progress method for emotion recognition that allows to exploit facial landmarks. Although the results computed on the JAFFE dataset show that there is still place for improvement, we believe that this approach has a strong potential to outperform currently proposed methods. In future work, we will therefore focus on improving our method by using attention mechanism on facial landmarks and experiment with additional loss function terms. We also plan to investigate other applications of our method, {\it e.g.} in the context of autistic children with incapabilities related to emotion recognition.

\begin{table}
\begin{center}
\begin{tabular}{|l|ccc|}
\hline
 & CK + & JAFFE & ISED \\
\hline\hline
CNN (2)  & 0.628  &   0.484   &  0.516  \\
CNN (5) &  0.728  &  \textbf{0.502}  &  0.593 \\
Inception-V3 &  0.304   &   0.268    & 0.479 \\
EmotionNet 2 & 0.204   &   0.249   & 0.21\\
EmotionalDAN & \textbf{0.736}  &  0.465 &  \textbf{0.62} \\
\hline
\end{tabular}
\end{center}
\caption{Cross-database accuracy results compared for different model architectures and seven emotion categories. All models are trained on AffectNet database. Face detection is applied as a preprocessing step on all test sets for all methods.}
\label{results-accuracy-7}
\end{table}

\begin{table}
\begin{center}
\begin{tabular}{|l|ccc|}
\hline
 & CK + & JAFFE & ISED \\
\hline\hline
CNN (2)     &  0.819   &  0.525   &  0.814  \\
CNN (5)  &  0.92  & \textbf{0.765}  & 0.867\\
Inception-V3 &  0.582  &  0.536  & 0.673 \\
EmotionNet 2 & 0.478   &   0.497   & 0.587\\
EmotionalDAN & \textbf{0.921} & 0.634 & \textbf{0.896} \\
\hline
\end{tabular}
\end{center}
\caption{Cross-database accuracy results compared for different model architectures and three emotion categories - positive, negative and neutral.}
\label{results-accuracy-3}
\end{table}

%
\bibliographystyle{ieeetr}
\bibliography{bibliography.bib}

\end{document}